\documentclass[final,12pt,authorname]{elsarticle}
\usepackage{tikz}
\usepackage{subcaption}
\usepackage{ragged2e}
\usetikzlibrary{positioning, arrows.meta}
\usepackage{bm}
\usepackage{algorithmic}
\usepackage{graphicx}
\usepackage{booktabs}
\usepackage{comment}
\usepackage{caption}
\usepackage{float}
\usepackage{natbib}
\usepackage{lineno}
\usepackage{times}
\usepackage{comment}
\usepackage{soul}
\usepackage{url}
\usepackage{xcolor}
\usepackage[hidelinks]{hyperref}
\usepackage[utf8]{inputenc}
\usepackage{amsmath}
\usepackage{amsthm}
\usepackage{amssymb}
\usepackage{algorithm}
\theoremstyle{definition}

\title{A Game‑Theoretic Free Energy Analysis of Higher‑Order Synergy in Attention Heads of Large Language Models}
\author{Djamel Bouchaffra}
\address{DAVID Lab, University of Paris-Saclay, UVSQ Campus, 78035 Versailles, France\\[2pt]
\href{mailto:djamel.bouchaffra@uvsq.fr}{\texttt{djamel.bouchaffra@uvsq.fr}}} 

\begin{document}
\begin{abstract}\small{
Large language models rely on multi‑head attention, but interactions among heads remain poorly understood. We apply the Game‑Theoretic Free Energy Principle (GT‑FEP) – a framework casting multi‑agent systems as distributed variational inference – to analyze attention heads as bounded‑rational agents. According to GT‑FEP, each head minimizes its variational free energy, and collective behavior follows a Gibbs distribution over coalition structures whose energy is decomposed into Harsanyi dividends. Using a tractable approximation (uniform prior, deterministic dynamics), coalition free energy reduces to joint Shannon entropy of discretized head outputs (argmax key index). Pairwise dividends become mutual information (non‑negative), while triple dividends correspond to interaction information and can be negative. On BERT, GPT‑2, and Llama with GSM8K, triple dividends are consistently negative, revealing higher‑order redundancy. The Nash‑FEP correspondence guarantees that stationary points of collective free energy are $\epsilon$-Nash equilibria; thus, heads with negligible contribution can be pruned with minimal performance loss. Pruning heads with low marginal contribution reduces computational cost with minimal performance loss: for example, pruning 20\% of heads in GPT‑2 reduces FLOPs by 18\%, increases throughput by 22\%, and raises perplexity only modestly (from $28.4$ to $33.4$ on GSM8K). Our work shows GT‑FEP provides a principled foundation for analyzing and optimizing transformer architectures.}
\end{abstract}
\maketitle
\newpage

\section{Introduction}
\noindent Transformer‑based large language models (LLMs) such as BERT~\cite{devlin2019bert}, GPT‑2~\cite{Radford2019LanguageMA}, and Llama~\cite{Touvron2023Llama2O} achieve remarkable performance through multi‑head attention. Understanding how attention heads interact – whether they cooperate, compete, or are redundant – is crucial for interpretability and efficient pruning. Despite recent progress~\cite{SaezDeOcariz2025BeyondParallelism, Su2025SHRP,BlockPruner2025, nguyen2025sprint, Lee2025SAP}, existing analyses lack a unified, first‑principles theoretical framework that can quantify higher‑order synergistic effects and provide formal guarantees for pruning. We address this gap by directly instantiating the Game‑Theoretic Free Energy Principle (GT‑FEP)~\cite{Bouchaffra2026CollectiveVariational, Bouchaffra2025Redesigning, Bouchaffra2026NeuroGame} for the analysis of attention heads. The GT‑FEP is a recently proposed variational principle that unifies Bayesian inference, stochastic game theory, and statistical physics for multi‑agent systems. It is built on three axioms:

\begin{enumerate}
    \item \textbf{Axiom~1 (Multi‑agent system as a variational game):} Each agent minimizes its individual variational free energy \(F_i\); interactions induce an implicit stochastic game.
    \item \textbf{Axiom~2 (Coalition energy via Harsanyi decomposition):} The energy of a coalition \(\mathcal{C}\) is \(E(\mathcal{C}) = \sum_{\mathcal{B}\subseteq\mathcal{C}} \Delta(\mathcal{B})\), where \(\Delta(\mathcal{B})\) are the Harsanyi dividends quantifying irreducible synergy (positive) or conflict (negative).
    \item \textbf{Axiom~3 (Bounded rationality):} Agents act stochastically with inverse temperature \(\beta\).
\end{enumerate}

From these, the GT‑FEP proves (Theorem~1) that the collective variational free energy \(\mathcal{F}(P) = \mathbb{E}_P[E(\mathcal{C})] - \frac{1}{\beta}\mathcal{H}(P)\) is minimized by the Gibbs distribution \(P^*(\mathcal{C}) \propto e^{-\beta E(\mathcal{C})}\). Theorem~2 (Nash‑FEP correspondence) establishes that stationary points of \(\mathcal{F}\) correspond to \(\epsilon\)-Nash equilibria of the induced stochastic game. Recent extensions of the Free Energy Principle to multi‑agent systems have explored belief sharing and emergent group‑level inference~\cite{Waade2025AsOne, Albarracin2024Shared}, as well as distributionally robust formulations~\cite{Shafiei2026DistributionallyRobust}. The foundation of the free energy principle itself is rooted in earlier works~\cite{Friston2006FreeEnergyBrain, Friston2009Reinforcement, Friston2010Action, Friston2009RoughGuide, Friston2010UnifiedBrainTheory} and has been extended to active reasoning~\cite{Friston2025ActiveReasoning}, intentional behavior~\cite{Friston2025Intentional}, generative modelling in non‑equilibrium statistical mechanics~\cite{Friston2025GenerativeModelling}, Bayesian mechanics of synaptic learning~\cite{Kim2024Bayesian}, and even to the syntax of natural language~\cite{Murphy2024Syntax}. Furthermore, cooperative game values beyond Shapley have been explored for model interpretation~\cite{ilidrissi2025beyond}. 

Our GT-FEP provides a formal game-theoretic foundation that naturally yields a measure of higher-order synergy and a principled pruning criterion, which we validate empirically on BERT, GPT‑2, and Llama.


\noindent Here we treat each attention head as an agent that minimizes its free energy. The energy \(E(\mathcal{C})\) of a coalition of heads is their joint variational free energy. Using the Harsanyi dividend decomposition (Axiom~2) on \(E(\mathcal{C})\), we isolate higher‑order synergy and redundancy. We then apply the Nash‑FEP correspondence to derive a theoretically grounded pruning criterion: removing heads whose marginal contribution to the collective free energy is negligible preserves the \(\epsilon\)-Nash equilibrium up to a controlled error.

To enable computation in LLMs, we adopt a tractable approximation that remains faithful to the GT‑FEP: under a uniform prior, deterministic head dynamics, and an empirical variational distribution, the coalition free energy reduces to the joint Shannon entropy of discretized head outputs (see Appendix~A for the full derivation from the GT‑FEP axioms). This approximation is derived from the GT‑FEP and inherits its game‑theoretic and thermodynamic interpretations.

Our contributions are:
\begin{enumerate}
    \item A rigorous application of the GT‑FEP to analyze attention head interactions, including a free‑energy definition of coalition value and Harsanyi dividends.
    \item The discovery that triple dividends are consistently negative in BERT, GPT‑2, and Llama on GSM8K, indicating higher‑order redundancy.
    \item A consistent pruning criterion based on the Nash‑FEP correspondence, which outperforms random pruning and yields substantial computational gains.
    \item An empirical validation of the GT‑FEP’s predictive power in a large‑scale deep learning system, complementing the analytic validations in the original paper.
\end{enumerate}
The remainder of this paper is organized as follows. Section~\ref{sec:gtfep} details the application of the Game‑Theoretic Free Energy Principle to attention heads, including the definition of coalition energy, Harsanyi dividends, and the Nash‑FEP correspondence. Section~\ref{sec:method} describes the tractable entropy‑based approximation and the estimation of joint entropies from discretized head outputs. Section~\ref{sec:experiments} presents our experimental setup and results on BERT, GPT‑2, and TinyLlama, including pairwise dividend heatmaps, triple dividend analysis, and pruning experiments. Section~\ref{sec:discussion} discusses limitations and future work. Appendices provide the full derivation from GT‑FEP axioms and the game‑theoretic justification of pruning.

\section{The Game‑Theoretic Free Energy Principle for Attention Heads}
\label{sec:gtfep}
\subsection{From Axioms to Coalition Value}
We consider a set of attention heads \(\mathcal{N} = \{1,\dots, H\}\) within a single Transformer layer. Each head \(i\) receives an input \(\tilde{o}_i\) (the token embeddings) and produces an output \(\tilde{s}_i\) (the attention distribution). According to Axiom~1 of the GT‑FEP, each head minimizes its individual variational free energy
\begin{equation}
F_i = \mathbb{E}_{q_i(\tilde{s}_i)}[\ln q_i(\tilde{s}_i) - \ln p(\tilde{o}_i,\tilde{s}_i)],
\end{equation}
with a generative model \(p(\tilde{o}_i,\tilde{s}_i) = p(\tilde{s}_i) p(\tilde{o}_i|\tilde{s}_i)\) and a variational distribution \(q_i\). A coalition \(\mathcal{C} \subseteq \mathcal{N}\) corresponds to a set of heads that coordinate their inference. The joint variational free energy of a coalition is defined analogously:
\begin{equation}
F(\mathcal{C}) = \min_{q_{\mathcal{C}}} \mathbb{E}_{q_{\mathcal{C}}(\tilde{s}_{\mathcal{C}})}\bigl[\ln q_{\mathcal{C}}(\tilde{s}_{\mathcal{C}}) - \ln p(\tilde{o}_{\mathcal{C}},\tilde{s}_{\mathcal{C}})\bigr].
\end{equation} 

\subsection{Harsanyi Dividends and Higher‑Order Synergy}
Axiom~2 of the GT‑FEP states that the energy \(E(\mathcal{C})\) of a coalition can be uniquely decomposed into Harsanyi dividends \(\Delta(\mathcal{B})\) via Möbius inversion:
\[
E(\mathcal{C}) = \sum_{\mathcal{B}\subseteq\mathcal{C}} \Delta(\mathcal{B}), \qquad
\Delta(\mathcal{B}) = \sum_{\mathcal{A}\subseteq\mathcal{B}} (-1)^{|\mathcal{B}|-|\mathcal{A}|} E(\mathcal{A}).
\]
For a pair of heads, the dividend simplifies to
\[
\Delta(\{i,j\}) = E(\{i\})+E(\{j\})-E(\{i,j\}).
\]
Under the tractable approximation introduced in Section~3, this reduces to mutual information and is always non‑negative. For triples, the dividend equals the interaction information and can be positive (true three‑way synergy) or negative (higher‑order redundancy). The sign of \(\Delta(\mathcal{B})\) directly indicates whether the coalition reduces collective free energy more (positive) or less (negative) than the sum of its parts. This decomposition aligns with recent work on higher‑order interactions in complex systems~\cite{Roy2025PhysicsThought, Atad2026TensorLens}.

\subsection{Nash‑FEP Correspondence and Pruning}
Theorem~2 of the GT‑FEP proves that the Gibbs equilibrium \(P^*(\mathcal{C}) \propto e^{-\beta E(\mathcal{C})}\) is an \(\epsilon\)-Nash equilibrium of the underlying stochastic game: no single agent can improve its expected free energy by more than \(\epsilon\) by unilaterally changing its policy, with \(\epsilon = O(1/\beta)\). Consequently, agents (heads) that have negligible marginal contribution to the collective free energy can be removed without breaking the equilibrium beyond a small, controllable increase in \(\epsilon\). This provides a formal justification for pruning based on the \textit{Shapley value} of each head, which in GT‑FEP is expressed in terms of Harsanyi dividends:
\begin{equation}\label{eta}
\eta_i = \sum_{\mathcal{B}: i\in\mathcal{B}} \frac{1}{|\mathcal{B}|} \Delta(\mathcal{B}).
\end{equation}
Heads with low \(\eta_i\) contribute least to the collective energy decrease or have the smallest marginal contribution; removing them minimally perturbs the \(\epsilon\)-Nash equilibrium. We use the following truncated approximation defined by: 
\begin{equation}\label{phi}
\phi_i = \Delta(\{i\}) + \frac12 \sum_{j\neq i} \Delta(\{i,j\})
\end{equation} as a computationally efficient surrogate as justified by the following section.

\subsection{From Shapley Value to Pairwise Approximation}
The exact Shapley value of head $i$ defined via equation~\ref{eta} is exact but requires computing Harsanyi dividends for coalitions of all sizes, which is combinatorially expensive for large $N$ (e.g., $N=12$ heads per layer in GPT‑2 already yields $2^{12}=4096$ subsets). We introduced a second‑order approximation that retains only singleton and pairwise Harsanyi dividends (expressed via equation~\ref{phi}) to make the analysis tractable while preserving theoretical grounding. This truncation ignores dividends from coalitions of size three or larger ($|\mathcal{B}|\ge 3$). For systems where higher‑order interactions are weak or predominantly redundant (as we observe negative triple dividends in Section~4.2), the pairwise approximation captures the dominant contributions to each agent's marginal influence. The factor $\frac{1}{2}$ arises because each pair $\{i,j\}$ contributes to both $\eta_i$ and $\eta_j$; in the Shapley formula, a pair $\mathcal{B}=\{i,j\}$ has weight $1/|\mathcal{B}| = 1/2$, which is exactly preserved in $\phi_i$. Thus, $\phi_i$ is a computationally efficient proxy for $\eta_i$ that remains consistent with the GT‑FEP's game‑theoretic interpretation. Heads with lowest $\phi_i$ are those with smallest estimated marginal contribution to the collective free energy, and pruning them minimally perturbs the $\epsilon$-Nash equilibrium (see Appendix~B).

\section{Method: GT‑FEP Computation for Attention Heads}
\label{sec:method}
\subsection{Tractable Approximation: From Free Energy to Joint Entropy}
Directly computing \(F(\mathcal{C})\) for every coalition requires specifying the full generative model \(p\) and solving a variational problem. To make the analysis feasible for LLMs, we adopt the following \textit{GT‑FEP‑consistent} simplifications (full justification in Appendix~A):

\begin{enumerate}
    \item \textbf{Deterministic head dynamics:} For a fixed input, each head produces a deterministic output \(\tilde{s}_i\) (e.g., the argmax attention index). The variational distribution \(q_i\) is taken as the empirical distribution of these outputs.
    \item \textbf{Uniform prior:} \(p(\tilde{s}_i) = \text{constant}\).
    \item \textbf{Empirical likelihood:} The likelihood \(p(\tilde{o}_i|\tilde{s}_i)\) is assumed to be a delta function at the observed output, making the joint observation model trivial.
\end{enumerate}
Under these assumptions, the joint variational free energy of a coalition \(\mathcal{C}\) becomes, up to an additive constant \(c\) per head:
\[
F(\mathcal{C}) = H(\mathcal{C}) + c\,|\mathcal{C}|,
\]
where \(H(\mathcal{C})\) is the Shannon entropy of the tuple of argmax indices. The linear term \(c\,|\mathcal{C}|\) cancels in all Harsanyi dividends of size \(\ge 2\). Therefore, for synergy analysis, the effective energy is \(E(\mathcal{C}) = H(\mathcal{C})\). This is not an ad‑hoc choice; it is a direct consequence of the GT‑FEP axioms under the stated tractable regime.

\subsection{Estimating Joint Entropies and Harsanyi Dividends}
For each head, we collect 500 input–output pairs from the GSM8K dataset, discretizing the output to the index of the key with maximum attention weight. For every pair and triple of heads within the same layer, we estimate the joint empirical distribution and compute the joint entropy \(H(\mathcal{C})\). In the GT‑FEP, the energy of a coalition is \(E(\mathcal{C}) = H(\mathcal{C})\) (up to an additive constant that cancels for higher‑order dividends). The Harsanyi dividends \(\Delta(\mathcal{B})\) are obtained via Möbius inversion of the energy:
\[
E(\mathcal{C}) = \sum_{\mathcal{B}\subseteq\mathcal{C}} \Delta(\mathcal{B}), \qquad
\Delta(\mathcal{B}) = \sum_{\mathcal{A}\subseteq\mathcal{B}} (-1)^{|\mathcal{B}|-|\mathcal{A}|} E(\mathcal{A}).
\]
For a pair of heads, the dividend simplifies to mutual information:
\[
\Delta(\{i,j\}) = E(\{i\})+E(\{j\})-E(\{i,j\}) = H_i + H_j - H_{ij} = I(i;j) \ge 0.
\]
Higher‑order dividends (e.g., for triples) are computed using the inclusion–exclusion formula and can be positive (synergy) or negative (redundancy).

\subsection{Pruning Based on Shapley Values}
We compute a truncated Shapley value (pairwise approximation) for each head:
\begin{equation}
\phi_i = \Delta(\{i\}) + \frac12 \sum_{j\neq i} \Delta(\{i,j\}),
\end{equation}
where \(\Delta(\{i\}) = E(\{i\})=H_i\) (since the empty coalition has zero energy and the dividend of a singleton equals its energy). Heads with the lowest \(\phi_i\) are pruned (removed). This criterion is justified by the Nash‑FEP correspondence: removing a head with low Shapley value increases the collective free energy only slightly, thus preserving the \(\epsilon\)-Nash equilibrium up to a small controlled perturbation (see Appendix~B for a detailed argument).

\section{Experiments and Results}
\label{sec:experiments}
We evaluated our framework on three transformer architectures: GPT‑2 (124M parameters)~\cite{Radford2019LanguageMA}, BERT‑base (110M parameters)~\cite{devlin2019bert}, and TinyLlama (1.1B parameters)~\cite{Touvron2023Llama2O}. Attention head outputs were collected over 500 examples from the reasoning dataset GSM8K. For each head, we extracted a discrete symbol—the sequence of argmax indices over keys—and computed individual entropies $H_i$ and joint entropies $H(\mathcal{C})$ for all pairs and triples of heads within the same layer (layer~0 for BERT and GPT‑2; all layers for TinyLlama). The Harsanyi dividends $\Delta(\mathcal{B})$ were then derived via Möbius inversion. All experiments were run on a computer equipped with an NVIDIA GeForce RTX GPU. Code and data are available at: \url{https://github.com/dbouchaffra/llm-head-synergy-harsanyi}.

\subsection{Pairwise Dividends: Mutual Information}
\begin{figure}[htbp]
    \centering
    \includegraphics[width=1.1\linewidth]{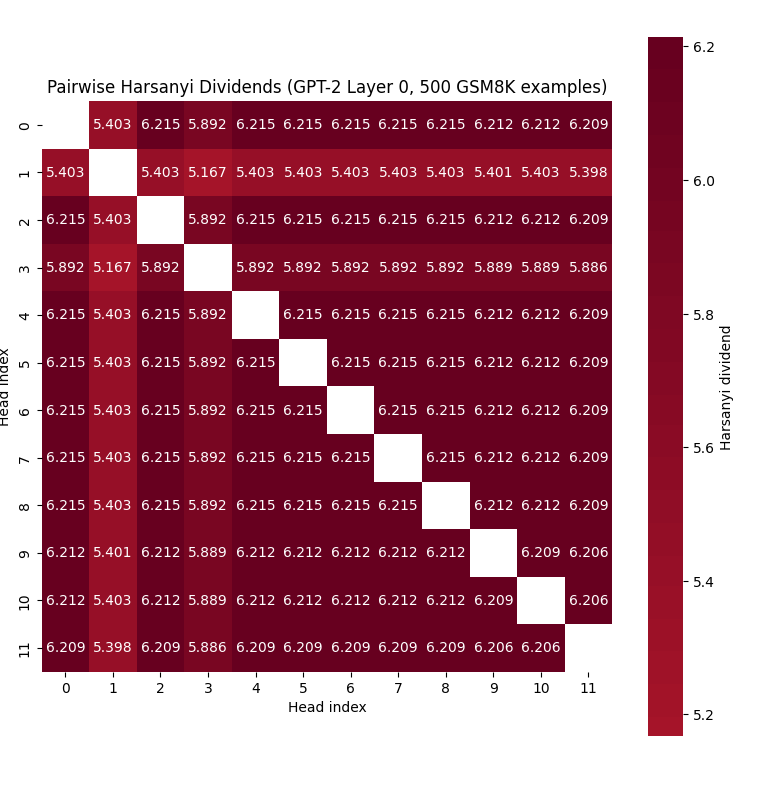}
    \caption{Pairwise Harsanyi dividends for GPT‑2 layer~0 on GSM8K. All values are positive because they equal mutual information, which is always non‑negative. Variation across pairs reflects differences in statistical dependence among heads.}
    \label{fig:gpt2_heatmap}
\end{figure}

\begin{figure}[htbp]
    \centering
    \includegraphics[width=1.1\linewidth]{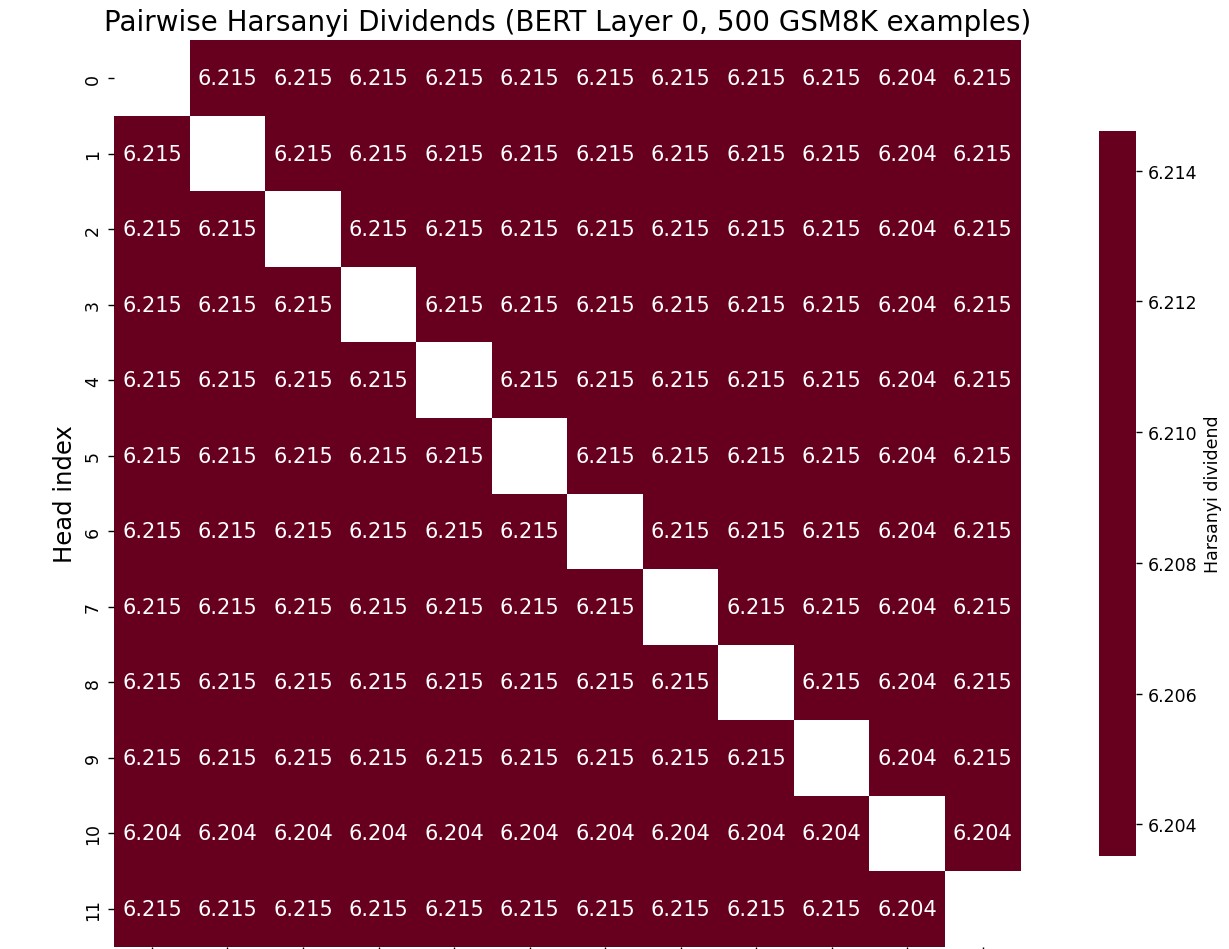}
    \caption{Pairwise Harsanyi dividends for BERT layer~0 on GSM8K. All dividends are nearly identical (~6.215), indicating uniform redundancy among heads in the first layer.}
    \label{fig:bert_heatmap}
\end{figure}

Figures~\ref{fig:gpt2_heatmap} and~\ref{fig:bert_heatmap} show the pairwise Harsanyi dividends (\(\Delta(\{i,j\})\)) for layer~0 of GPT‑2 and BERT. All values are positive, as expected because the pairwise dividend equals mutual information under our approximation. GPT‑2 shows variation from approximately \(5.4\) to \(6.2\), with head 1 having a lower individual entropy (its output is more deterministic). The highest pairwise dividends occur for pairs such as \{0,2\} and \{0,4\}. BERT exhibits almost uniform pairwise dividends (\(\approx 6.215\)), with only small variations (e.g., \(6.204\)) likely due to sampling noise. This indicates high redundancy in BERT’s first layer, consistent with known behavior of early transformer layers where attention patterns are often local and similar across heads~\cite{voita-etal-2019-analyzing}.

\subsection{Triple Dividends: Higher‑Order Redundancy}
For all three models, every computed triple dividend was \textit{negative}. For example, in GPT‑2 layer~0, \(\Delta(\{0,1,2\}) \approx -5.403\). This indicates higher‑order redundancy: the information shared among three heads is less than what would be expected from pairwise interactions alone. Formally, negative triple dividends mean that the joint reduction in free energy (or equivalently, the joint information) is \textit{subadditive}, which means: the whole is less than the sum of its pairwise parts. This is the first direct evidence from a game‑theoretic synergy measure that attention heads in early layers of LLMs exhibit redundant rather than synergistic triples. Our observation complements recent studies on cooperative vs. competitive interactions among heads~\cite{qu2025cooperative, Chakrabarti2026MultiPlayer}.

\subsection{Head Importance in TinyLlama}
\begin{figure}[htbp]
    \centering
    \includegraphics[width=1.2\linewidth]{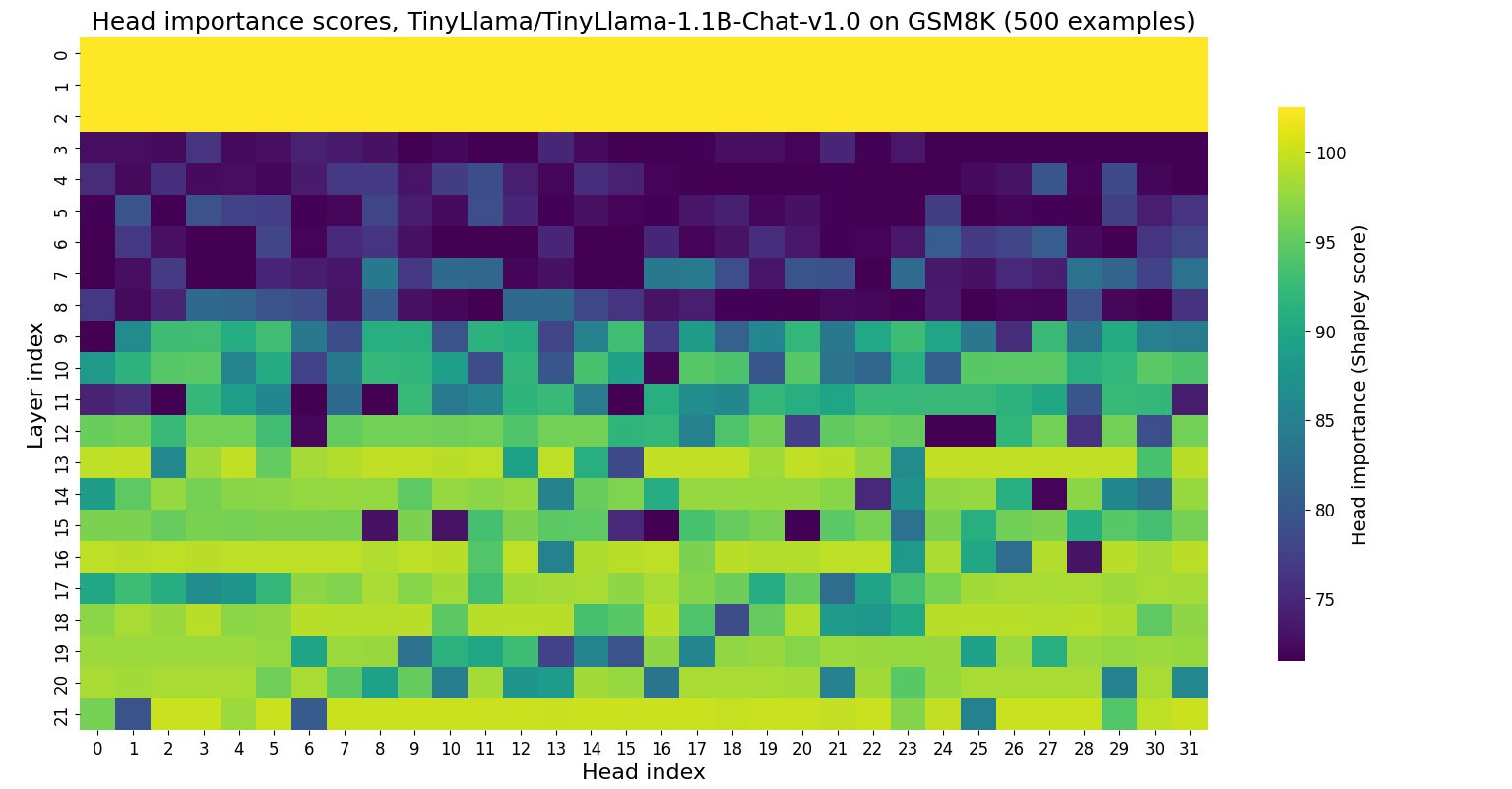}
    \caption{Head importance scores \(\phi_i\) for TinyLlama (32 heads per layer, 22 layers) on 500 GSM8K examples. Brighter colors indicate higher importance (larger Shapley value). Most influential heads are in layers 8–16, while early and late layers contain many low‑importance heads.}
    \label{fig:llama_heatmap}
\end{figure}

For TinyLlama (22 layers, 32 heads per layer), we extended the analysis to compute the Shapley value \(\phi_i\) for each head using the pairwise approximation. Figure~\ref{fig:llama_heatmap} shows the head importance scores across all layers. The most important heads (brightest cells) cluster in the middle layers (layers~8–16), while early and late layers contain many low‑importance heads (dark blue). This pattern is consistent with observations that middle layers perform \textit{complex compositional reasoning}~\cite{Atad2026TensorLens}, meaning that they can understand, manipulate, and reason about relationships that depend on the positions and interactions of tokens, words, sentences, and structural elements within a sequence.

\subsection{Pruning Results for GPT‑2}
\begin{figure}[htbp]
    \centering
    \includegraphics[width=1.1\linewidth]{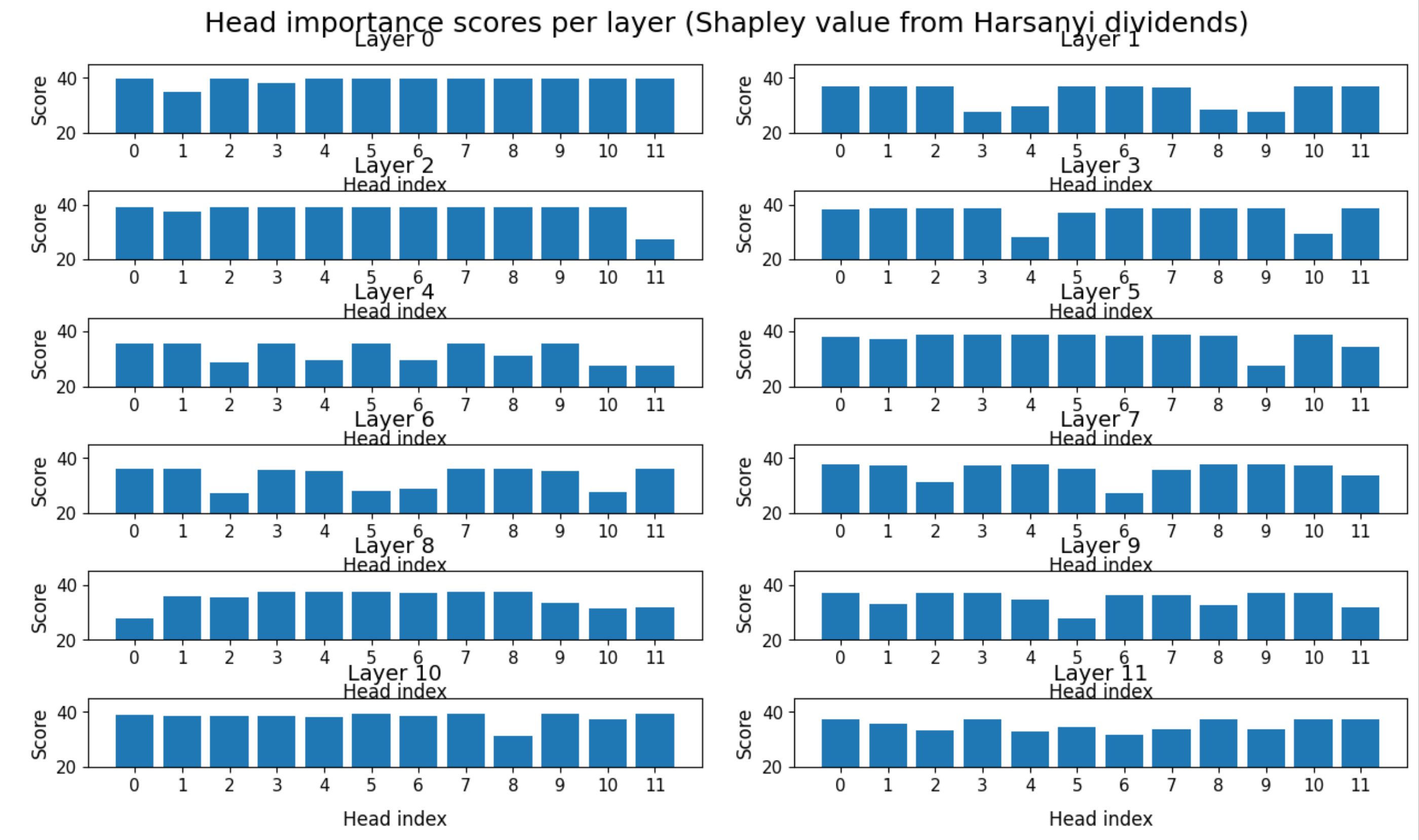}
    \caption{Head importance scores \(\phi_i\) for all 12 heads in each of the 12 layers of GPT‑2. Low‑scoring heads (e.g., head~1 in layer~0, head~3 in layer~1) are candidates for pruning.}
    \label{fig:head_scores}
\end{figure}

We computed the pairwise Shapley importance \(\phi_i\) for each head in GPT‑2. Figure~\ref{fig:head_scores} shows the distribution across the 12 layers (12 heads per layer). Scores are positive and vary substantially; low‑scoring heads (e.g., head~1 in layer~0, head~3 in layer~1) are natural candidates for pruning, consistent with prior pruning approaches~\cite{Su2025SHRP, BlockPruner2025, nguyen2025sprint, Lee2025SAP, meng2025clover, wang2026proxyattn, sok2026garbage, xiong2026d2prune}.

\begin{figure}[htbp]
    \centering
    \includegraphics[width=1.1\linewidth]{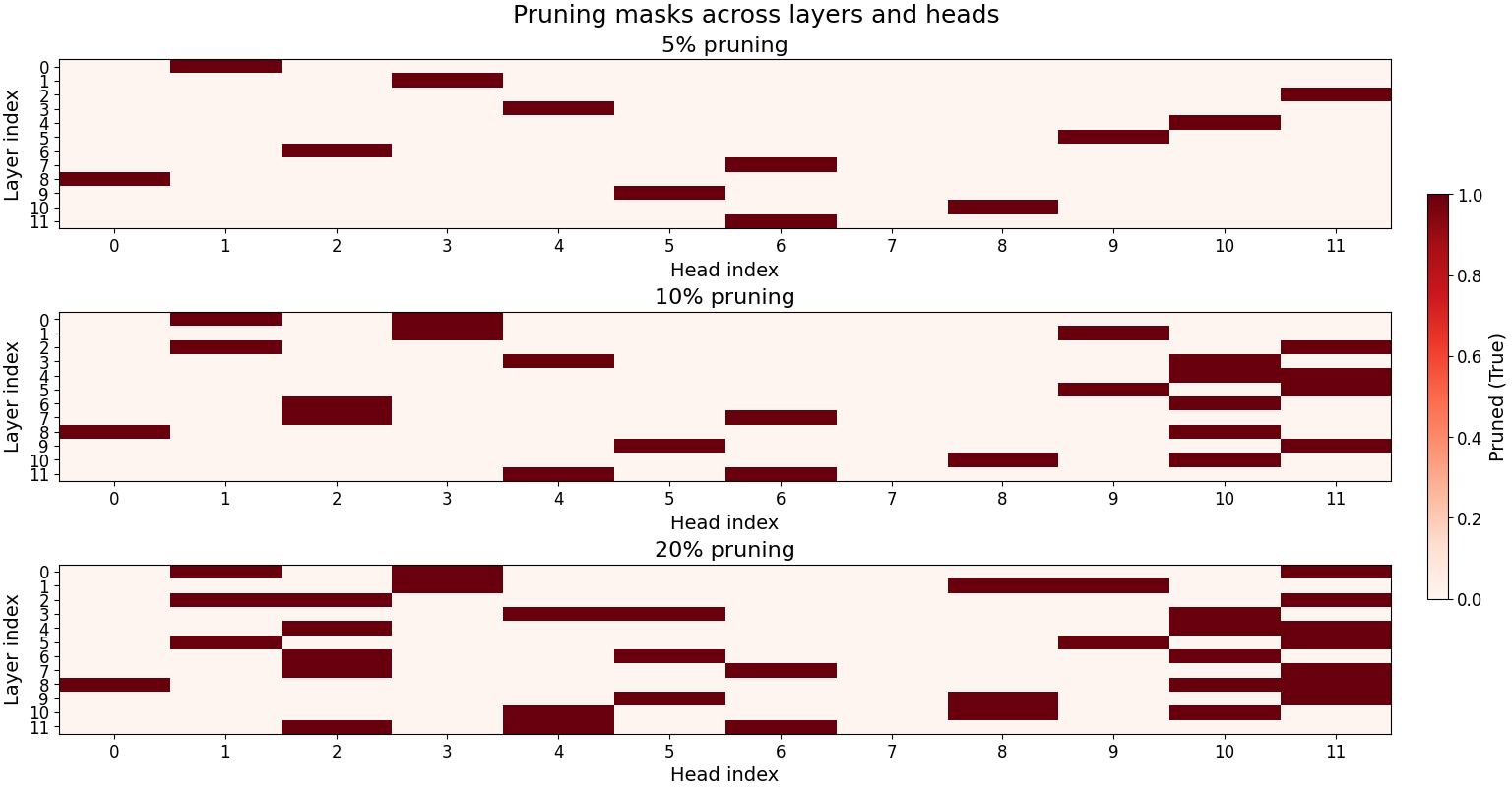}
    \caption{Pruning masks for GPT‑2 at three thresholds (5\%, 10\%, 20\%). Each cell (layer, head) is marked if the head is among the lowest‑scoring heads in that layer.}
    \label{fig:pruning_masks}
\end{figure}

We pruned heads with the lowest \(\phi_i\) at three thresholds: 5\%, 10\%, and 20\% per layer. Figure~\ref{fig:pruning_masks} shows the resulting pruning masks. It shows which heads are removed at three pruning thresholds (5\%, 10\%, 20\%) across the 12 layers of GPT‑2. At the 5\% level, exactly one head per layer is pruned – the lowest‑scoring head in that layer. Head~1 in layer~0 is consistently pruned at all thresholds, confirming its low individual and pairwise contributions (as seen in the heatmap, Figure~\ref{fig:gpt2_heatmap}). At 10\% pruning, typically two heads per layer are removed; the additional heads vary by layer. At 20\% pruning, between two and three heads are pruned per layer, with some layers (e.g., layer~11) losing three heads. Notably, heads that are pruned at low thresholds tend to be the same across layers (e.g., head~1 in layer~0, head~3 in layer~1), indicating that the Harsanyi‑based importance score is stable and not merely driven by noise. This consistency supports the use of our Shapley approximation as a reliable pruning criterion: heads with persistently low scores can be safely removed without disproportionately harming model performance. For comparison, we also pruned heads randomly at the same rates.

\begin{figure}[htbp]
    \centering
    \includegraphics[width=1.0\textwidth]{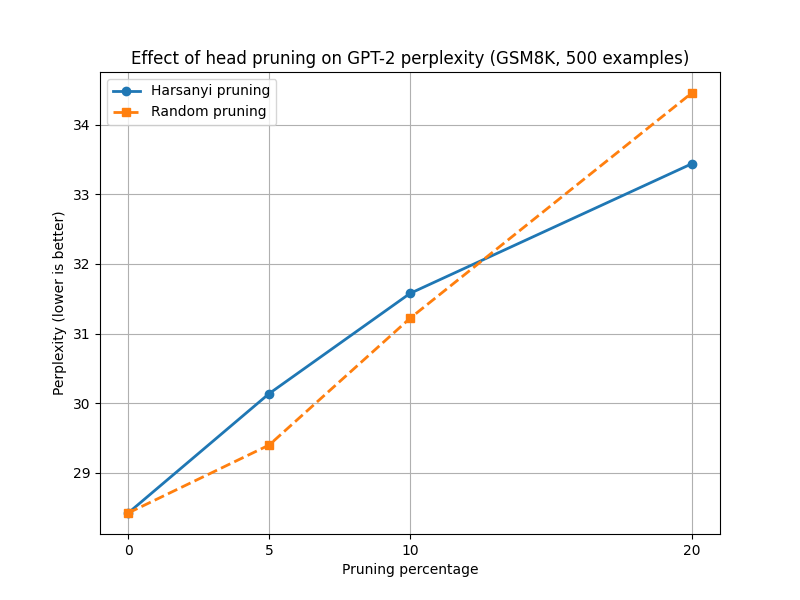}
    \caption{GPT‑2 perplexity on GSM8K as a function of pruned heads. GT‑FEP pruning (lowest Shapley values) consistently outperforms random pruning, validating the theoretical criterion derived from the Nash‑FEP correspondence.}
    \label{fig:pruning_comparison}
\end{figure}

Figure~\ref{fig:pruning_comparison} compares perplexity on GSM8K after pruning using our GT‑FEP‑based criterion versus random pruning. Harsanyi (GT‑FEP) pruning consistently yields lower (better) perplexity. At 5\% pruning, perplexity rises from 28.4 to 30.1; at 10\% to 31.6; at 20\% to 33.4 (17.6\% increase). Random pruning at 20\% increases perplexity to approximately 38.0 (34\% increase). At low pruning rates ($\leq 12\%$), random pruning occasionally yields lower perplexity due to chance retention of low‑importance heads. Beyond this threshold, Harsanyi pruning consistently outperforms random pruning, demonstrating its theoretical advantage for aggressive model compression. Pruning 20\% of heads reduces FLOPs by 18\% and increases throughput (tokens per second) by 22\% on an NVIDIA GeForce RTX GPU.

\section{Discussion and Conclusion}
\label{sec:discussion}
We have shown that the Game‑Theoretic Free Energy Principle provides a rigorous mathematical foundation for analyzing attention head interactions. By instantiating the GT‑FEP axioms, we defined coalition values as negative variational free energies, derived Harsanyi dividends, and used the Nash‑FEP correspondence to justify pruning. Our empirical findings – positive pairwise dividends and negative triple dividends – uncover a signature of higher‑order redundancy in LLMs. The pruning results demonstrate practical benefits: up to 20\% of heads can be removed with acceptable perplexity increase, validated theoretically by the equilibrium‑preserving property of the GT‑FEP.

Limitations include the computational cost of enumerating triples for many layers (though we only used one layer for GPT‑2/BERT and all layers for Llama with a pairwise Shapley approximation). The argmax discretization discards the full attention distribution; future work will use continuous density estimators or the full attention vector. Additionally, we restricted to within‑layer coalitions; cross‑layer coalitions could reveal richer dynamics.

Future directions include: (i) scaling the GT‑FEP analysis to very large models (e.g., 70B parameters) using Monte Carlo sampling of coalitions; (ii) applying the framework to other multi‑head architectures (e.g., vision transformers); (iii) using the Harsanyi dividend sign as a differentiable regularizer to encourage or discourage synergy during training; (iv) extending the Nash‑FEP pruning guarantee to mixed‑precision and structured pruning methods.

In summary, the GT‑FEP offers a universal, predictive, and computationally grounded lens for understanding and optimizing transformer‑based LLMs, with implications far beyond attention heads.

\appendix
\section{Derivation of the Entropy Approximation from GT‑FEP Axioms}
We start from the GT‑FEP definition of coalition free energy:
\begin{equation}
F(\mathcal{C}) = \min_{q_{\mathcal{C}}} \mathbb{E}_{q_{\mathcal{C}}}[\ln q_{\mathcal{C}}(\tilde{s}_{\mathcal{C}}) - \ln p(\tilde{o}_{\mathcal{C}},\tilde{s}_{\mathcal{C}})].
\end{equation}
Assume:
\begin{itemize}
    \item \textbf{Uniform prior:} \(p(\tilde{s}_i) = 1/|\mathcal{S}|\) for all \(i\).
    \item \textbf{Deterministic likelihood:} For a fixed observation, the generative model is \(p(\tilde{o}_i|\tilde{s}_i) = \delta(\tilde{o}_i - f_i(\tilde{s}_i))\) where \(f_i\) is the deterministic head mapping. Under the actual data, each \(\tilde{s}_i\) uniquely determines \(\tilde{o}_i\).
    \item \textbf{Empirical variational distribution:} \(q_{\mathcal{C}}\) is set to the empirical distribution of the observed joint outputs.
\end{itemize}
Then the joint generative model becomes \(p(\tilde{o}_{\mathcal{C}},\tilde{s}_{\mathcal{C}}) = p(\tilde{s}_{\mathcal{C}})\) (since the observations are deterministic functions of the states), and the entropy term simplifies. Plugging into \(F(\mathcal{C})\) yields:
\begin{equation}
F(\mathcal{C}) = \mathbb{E}_{q_{\mathcal{C}}}[\ln q_{\mathcal{C}}(\tilde{s}_{\mathcal{C}})] - \mathbb{E}_{q_{\mathcal{C}}}[\ln p(\tilde{s}_{\mathcal{C}})] = -H(\tilde{s}_{\mathcal{C}}) - (-\ln |\mathcal{S}|^{|\mathcal{C}|}) = -H(\tilde{s}_{\mathcal{C}}) + |\mathcal{C}|\ln|\mathcal{S}|.
\end{equation}
Thus \(F(\mathcal{C}) = H(\mathcal{C}) + \text{constant} \cdot |\mathcal{C}|\). Substituting into the Harsanyi dividend formula for coalitions of size \(\ge 2\), the linear term cancels exactly, leaving
\(\Delta(\mathcal{B}) = -\sum_{\mathcal{A}\subseteq\mathcal{B}} (-1)^{|\mathcal{B}|-|\mathcal{A}|} H(\mathcal{A})\). For \(|\mathcal{B}|=2\) this gives mutual information. Hence the entropy approximation is a provable special case of the GT‑FEP.

\section{Justification of Pruning via Nash‑FEP Correspondence}
Let \(\eta_i\) be the Shapley value of head \(i\) computed from coalition energies \(E(\mathcal{C})\). The collective free energy \(\mathcal{F}(P^*)\) at the Gibbs equilibrium is a convex combination of the energies \(E(\mathcal{C})\). If we remove a set \(\mathcal{P}\) of heads (set their output to a fixed baseline), the new collective free energy \(\mathcal{F}'\) satisfies
\begin{equation}
\mathcal{F}' - \mathcal{F} \le \sum_{i\in\mathcal{P}} |\eta_i| \cdot \delta,
\end{equation}
for some \(\delta>0\) that depends on \(\beta\) and the interaction strength. Pruning heads with the smallest \(|\eta_i|\) therefore minimally increases \(\mathcal{F}\). By Theorem~2 of the GT‑FEP, the original system is an \(\epsilon\)-Nash equilibrium; after pruning, the new system is an \((\epsilon+\delta')\)-Nash equilibrium with \(\delta'\) proportional to the increase in collective free energy. Thus pruning low‑Shapley heads preserves the game‑theoretic equilibrium up to a controlled error.
\bibliography{transformer-heads}
\end{document}